\documentclass{article}

\usepackage{PRIMEarxiv}
\usepackage{amsmath}
\usepackage{amssymb}
\usepackage{amsthm}
\usepackage[utf8]{inputenc} 
\usepackage[T1]{fontenc}    
\usepackage{hyperref}       
\usepackage{url}            
\usepackage{booktabs}       
\usepackage{amsfonts}       
\usepackage{nicefrac}       
\usepackage{microtype}      
\usepackage{lipsum}
\usepackage{fancyhdr}       
\usepackage{graphicx}       
\graphicspath{{media/}}     
\usepackage{booktabs}
\title{STLM Engineering Report:\\Dropout
}

\author{
  Dylan Hillier \\
  A*STAR and Singapore Management University  \\
  \texttt{das.hillier.2023@phdcs.smu.edu.sg} \\
  \And
  Leon Guertler \\
  A*STAR, Institute of High Performance Computing (IHPC)\\
  \texttt{guertlerlo@cfar.a-star.edu.sg} \\
   \And
  Bobby Cheng \\
  A*STAR, Institute of Infocomm Research (I2R) \\
  \texttt{bobby\_cheng@i2r.a-star.edu.sg} \\
  \And
  Cheston Tan \\
  A*STAR, Institute of High Performance Computing (IHPC) \\
  \texttt{cheston-tan@i2r.a-star.edu.sg} \\
}

\begin{document}
\maketitle

\begin{abstract}
In this work we explore the relevance of dropout for modern language models, particularly in the context of models on the scale of <100M parameters. We explore it's relevance firstly in the regime of improving the sample efficiency of models given small, high quality datasets, and secondly in the regime of improving the quality of its fit on larger datasets where models may underfit. 
We find that concordant with conventional wisdom, dropout remains effective in the overfitting scenario, and that furthermore it may have some relevance for improving the fit of models even in the case of excess data, as suggested by previous research. In the process we find that the existing explanation for the mechanism behind this performance gain is not applicable in the case of language modelling.
\end{abstract}


\section{Motivation}\label{sec:motivation}
In our earlier work, titled `Super Tiny Language Models'\cite{hillier2024supertinylanguagemodels}, we outlined our ambitions and approach for developing Super Tiny Language Models (STLMs) that are high performing in spite of having significantly smaller parameter counts. One of our listed strategies is experimentation over the effects of dropout.

First proposed by Srivastasa et al (2014)~\cite{srivastava2014dropout}, dropout involves the probabilistic disabling of neurons -- implemented by zeroing out the activations of a given layer. This is broadly equivalent to subsampling a smaller network, and then only performing gradient updates on this smaller network. As such a network trained with dropout is thought to be regularized through it's approximation of a larger ensemble of networks. While early uses of the technique had high dropout ratios of 0.5~\cite{srivastava2014dropout} -- dropping out half of the neurons on any given training pass, as data has been increasingly scaled, the dropout ratio has been scaled down, with models either using a ratio of 0.1, or for most modern large language models omitting it entirely~\cite{xue2024repeat}. For instance while early models like GPT-2~\cite{radford2019language} used dropout throughout the model (at a rate of 0.1, in addition to L2-Normalisation), recent works like Llama-2~\cite{touvron2023llama} don't use it at all.

\subsection{Sample Efficiency}
A large advantage of smaller language models, is that they can be trained with smaller amounts of data -- in principle this should allow models to be trained with higher quality data due to the increased ease of collecting such data at a scale sufficient for optimal performance in the sense of scaling laws~\cite{kaplan2020scaling}. Indeed highly-performant small models such as ORCA~\cite{mukherjee2023orca} and Phi~\cite{abdin2024phi} have relied on high quality synthetic data to outperform other models at their scale. Data efficient LLMs may also play a key role in empowering the modelling of under-resourced languages~\cite{costa2022no}.\\ One potential motivation for using dropout on these models then, is to increase this sample efficiency by enabling samples to be used over multiple epochs. Repeated training of LLMs across the data has been show to cause the phenomenon of a `Token-Crisis' - where multi-epoch training causes the performance of language models to degrade~\cite{xue2024repeat}. The promise of dropout then lies in its use as a regularization technique, and indeed it has been shown to be uniquely effective in this role for alleviating this token-crisis~\cite{xue2024repeat}, and is furthermore its standard usage across the field.  

\subsection{Reducing Underfitting}\label{sec: underfitting_expl}
In addition to being useful for combating overfitting and the ability to train with smaller datasets, Liu et al. (2024)~\cite{liu2023dropout}
report that dropout can also play a role in \textit{combatting underfitting}. The authors report that during the beginning of training the variance of mini-batch gradient directions and error in the gradient with respect to the full-batch gradient are respectively far higher. They argue that dropout can function to reduce this gradient variance and error during training and that it has the effect of preventing the model from overfitting to specific, high-norm batches.
Accordingly the authors propose using dropout in early stages to reduce the effect of this gradient variance. They demonstrate that across a variety of models and optimizers over imagenet, the performance of a model can be increased for underfitting models.




\section{Method}
\label{sec:method}
As described in Section~\ref{sec:motivation}, dropout with ratio $p$ consists in setting individual activations of a layer to $0$ with chance $p$. In particular for a given sample in the training data we define the dropout layer as $\text{Dropout}_p(\mathbf{x})$, given $\mathbf{x}:\mathbb{R}^d$, $\mathbf{m}:\mathbb{R}^d\sim \text{Bernoulli}(p)$ independently generated for each sample and layer:
\[
\text{Dropout}_p(\mathbf{x}) = \mathbf{m} * \mathbf{x}
\]Where $*$ is the elementwise multiplication operator.
In our experiments we apply dropout to the output of the embedding layers and the output of each attention layer. We find that placing the dropout in different parts of the model has minimal impact on the effects observed in section~\ref{sec: experiments}
\subsection{Dropout Schedulers}
While dropout itself is a simple enough concept, there may be some benefit to using a scheduler to mitigate any instability caused by switching the dropout on/off during training. Indeed in Liu et al.\cite{liu2023dropout} the authors experiment with a linear scheduled dropout. 

\begin{figure}[h]
    \centering
    \includegraphics[width=0.75\linewidth]{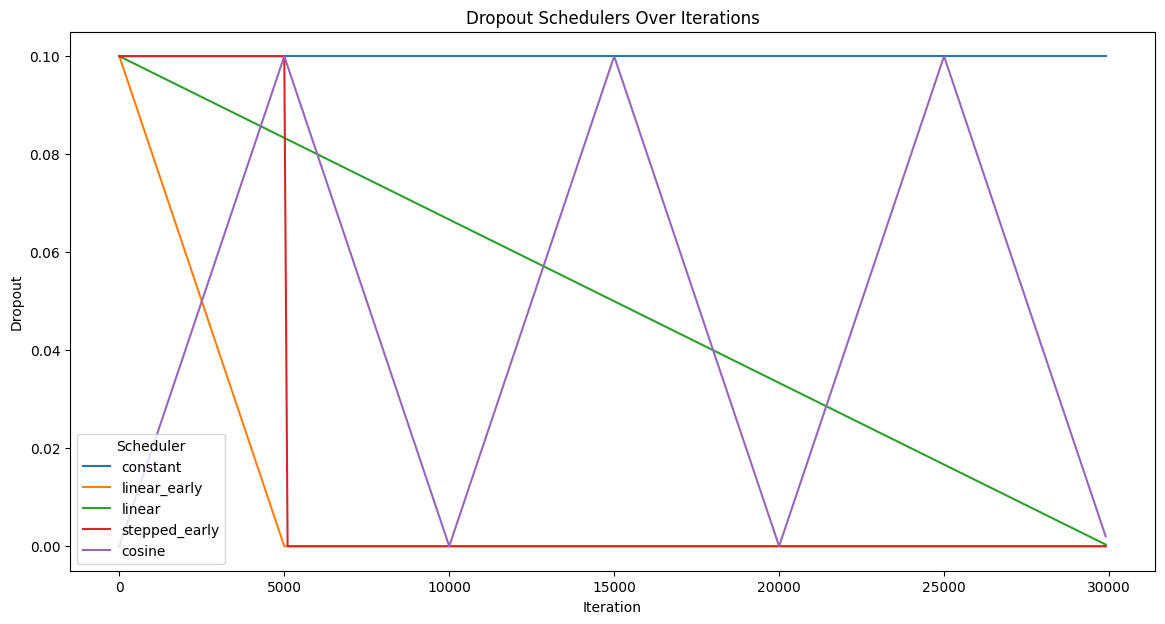}
    \caption{The dropout schedulers used in the underfitting experiment}
    \label{fig:underfitting}
\end{figure}

For the case of underfitting, we use the following schedulers, depicted in Figure~\ref{fig:underfitting}:
\begin{itemize}
    \item Constant dropout ratio of 0.1 (constant)
    \item Early dropout with a cutout after 5000 iterations (stepped early)
    \item Linear scheduled (decreasing) from 0.1 to 0 (linear) 
    \item Linear early schedule (decreasing) from 0.1 to 0 over 5000 iterations (linear early)
    \item Triangular Dropout with 3 cycles (triangular)
\end{itemize}

\begin{figure}[h]
    \centering
    \includegraphics[width=0.75\linewidth]{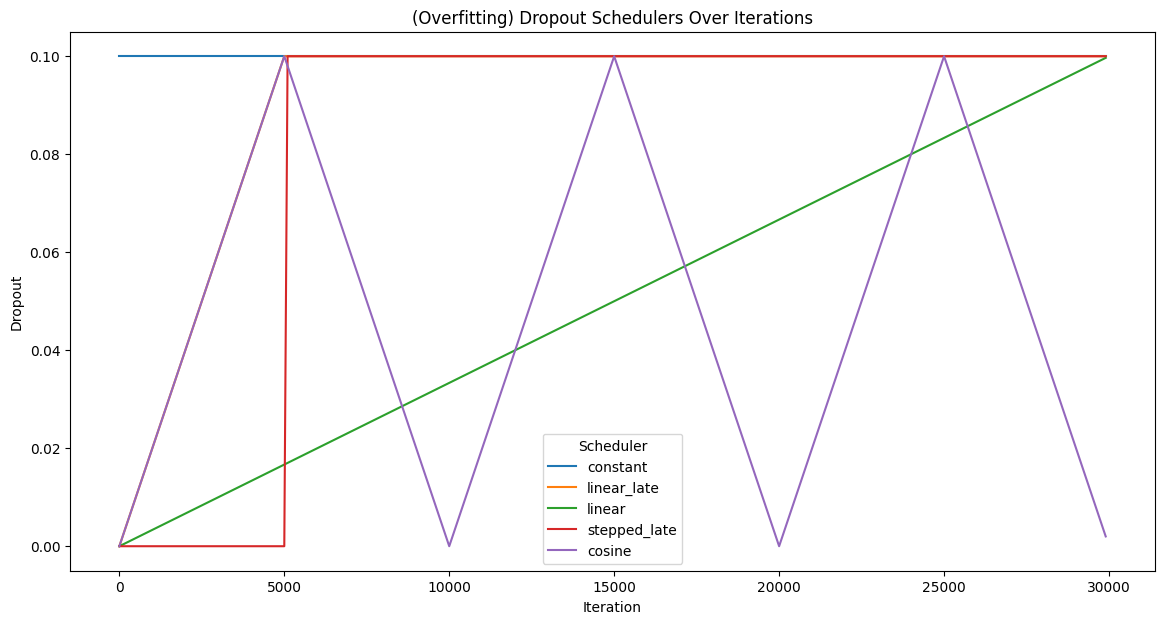}
    \caption{The Dropout schedulers used in the overfitting experiment}
    \label{fig:overfitting}
\end{figure}

For the case of overfitting, we use the following schedulers, depicted in Figure~\ref{fig:overfitting}:
\begin{itemize}
    \item Constant dropout ratio of 0/0.1 (constant)
    \item Linear schedule (increasing) from 0 to 0.1 (linear)
    \item Linear schedule (increasing) from 0 to 0.1 over 5000 iterations (linear\_late)
    \item Late dropout starting at 5000 iterations (stepped\_late)
    \item Triangular Dropout with 3 cycles (triangular)
\end{itemize}


\section{Experiments}\label{sec: experiments}
\begin{table}[h]
    \centering
    \small
    \begin{tabular}{lcccccc}
\toprule
Model & BLiMP & HellaSwag & ARC\_easy & WinoGrande & MMLU & In-Domain\\
\midrule
(OpenWebText) -- Underfitting & & & & & &\\
Constant Dropout (0) & 77\%& 29\%& 38\%& 50\%& 23\% & 2.29\\ 
Constant Dropout (0.1) & 78\% & 30\% & 38\%& 50\%& 24\% & 2.34\\
Triangular & 76\% & 30\% & 37\% & 51\% & 24\% & 2.33\\
\textbf{Linear Early (Decreasing)} & 77\% & 30\% & 39\% & 50\% & 25\% & \textbf{2.26}\\
Linear (Decreasing) & 77\% & 30\% & 39\% & 51\%& 24\% & 2.28\\
Hard Cutoff Early & 76\% & 29\% & 37\% & 51\% & 25\% & 2.33\\

\midrule
(Simple English Wikipedia) -- Overfitting& & & & & &\\
Constant Dropout (0) & 65\%& 28\%& 36\%& 50\%& 24\% & 2.26\\
Constant dropout  (0.1)& 69\%&28\%& 37\%& 49\%& 24\% & 1.88\\
Triangular & 68\% & 29\% & 36\% & 50\% & 24\% & 1.92\\
\textbf{Linear Early (Increasing)} & 68\% & 28\% & 38\% & 49\% & 24\% & \textbf{1.87}\\
Linear (Increasing) & 68\% & 28\% & 37\% & 50\% &23\% & 1.88\\
Hard Cutoff Early & 68\% & 28\% & 36\% & 50\% & 24\% & 1.88\\
\midrule
Metric  &Accuracy & Accuracy & Accuracy & Accuracy & Accuracy & Perplexity\\
Num. Choices                             & 2    & 4    & 4    & 2    & 4  & --  \\
Chance Perf.                            & 50\% & 25\% & 25\% & 50\% & 25\% & --\\
\bottomrule
\end{tabular}
    \caption{Results for models trained with different dropout schedulers across Simple English Wikipedia and OpenWebText respectively}
    \label{tab: exp res}
\end{table}
We report our results on the underfitting experiment in the first section of Table~\ref{tab: exp res}, where we can see that, while adding constant dropout decreases performance, early usages of dropout can improve performance, particularly using linear schedulers. The triangular and hard cutoff schedulers on the other hand appear to have hindered the performance of the model; in the case of the former this is in line with our expectation since this continues to play a normalizing role throughout the training.  While these findings do weakly endorse those of Liu et al. (2024)~\cite{liu2023dropout}, the performance gain is minor, and when we look at the distribution of different schedulers is largely centered around the performance of the baseline version with no dropout.

For the case of combatting overfitting we find little difference between dropout schedulers, albeit with  a slight improvement of performance in the case of the early linear increasing scheduler. In particular all the dropout techniques improve performance significantly over the baseline which we empirically observe begins overfitting after around 10000 epochs.
The one exception to this is that the triangular learning rate scheduler markedly decreases performance in comparison to the other schedulers.

More generally we can argue that across these two experiments, that schedules with annealed, continuous changes perform better than sharp discontinuous ones, presumably through avoiding instabilities in training.

\subsection{Mechanism Investigation}
\begin{figure}
    \centering
    \includegraphics[width=1\linewidth]{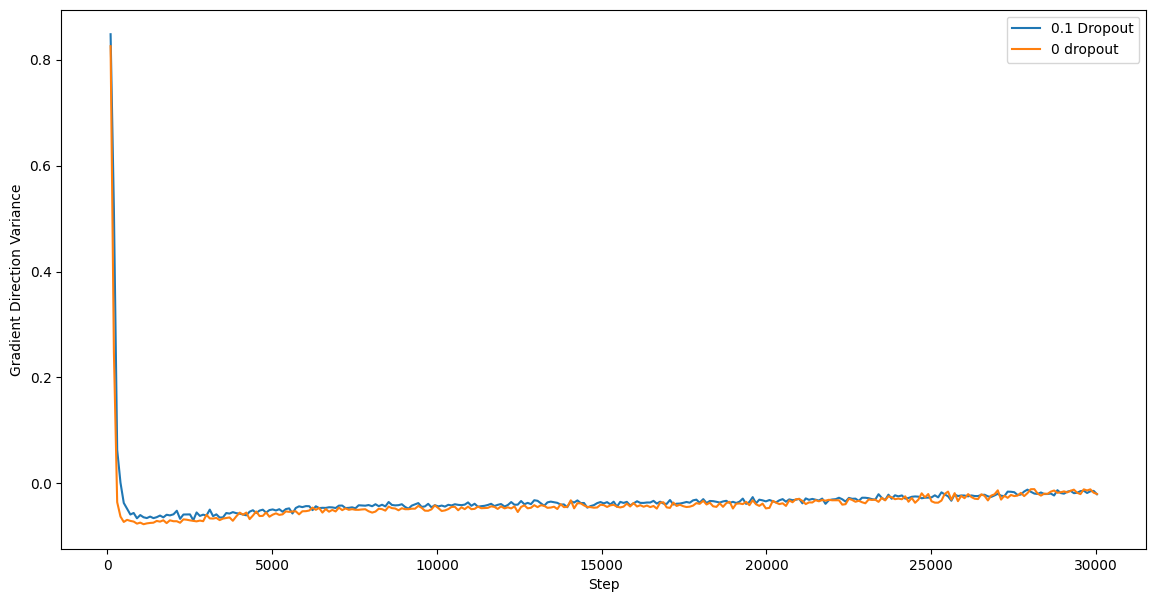}
    \caption{Gradient Direction Variance as a function of dropout and time}
    \label{fig: grad_var}
\end{figure}
As mentioned in Section~\ref{sec: underfitting_expl} Liu et al.~\cite{liu2023dropout} propose that the increased fit that arises with early dropout is as a result of it reducing the variance of gradient steps in the early stages of training. We initially hypothesised that this may be better explained by a curriculum effect: that the dropout simplifies the learning task during the early stages of training by reducing the expressivity of the model. As such we run a brief experiment using an annealed deterministic dropout (i.e. slowly introducing neurons over the early stages of training,) however it does not result in an improvement over the no-dropout baseline. We then try to investigate the variance mechanism, and thus record the gradient variance during training with/without dropout. Following Liu et al.~\cite{liu2023dropout} we calculate the following:
\[
\text{GDV}=\frac{1}{L\cdot\vert G\vert \cdot \vert G\vert -1} \cdot \sum_{l\in L,g_i,g_j\in G,i\neq j}<g^l_i,g^l_j>_{\text{cosine}}
\]
Where $G$ is a set of gradients produced over $10$ minibatches at a given fixed training checkpoint, and $L$ is the set of weights of a given model. \textit{Contrary} to the results reported by Liu et al. we find that the gradient variance is consistently higher throughout training when using dropout, as depicted in figure~\ref{fig: grad_var} . Indeed we find this is particularly true during the earlier stages of training, and that the divergence tapers off towards the end. It is not entirely surprising that dropout should increase the variance of gradient directions since it samples sub-networks, interrupting gradient flows in different ways across the batches.
\section{Conclusion}
We have identified two continued use cases for dropout in language modelling. In the first instance, we show that it remains useful on small datasets as a normalization technique to combat overfitting to the training data. Secondly we extend the findings of Liu et al. (2024)~\cite{liu2023dropout} to our language modelling task and reiterate their results: the judicious usage of dropout early on in the training process can result in an improved fit. In particular we find that regardless of whether it is for combating underfitting or overfitting, using an early linear schedule to transition between the use of dropout performs best. That said, the effect is relatively weak and we have not yet shown it to be robust to different choices of e.g. optimizer or across scales in the case of combatting underfitting.

We then perform a brief investigation into the mechanisms at play in this performance improvement, and find contrary evidence to the existing literature as to the role of dropout in reducing underfitting on our domain. As such we call for more investigation into the use of dropout for improving model performance along several axes:
\begin{itemize}
    \item Identifying suitable cutoff points and portions of training for which dropout is useful
    \item Identifying more plausible explanations for the mechanisms thereof
    \item Investigating how dropout and other regularization techniques interact with optimizers and learning rate schedulers.
\end{itemize}
\bibliographystyle{unsrt}  
\bibliography{references}  

\begin{thebibliography}{10}

\bibitem{hillier2024supertinylanguagemodels}
Dylan Hillier, Leon Guertler, Cheston Tan, Palaash Agrawal, Chen Ruirui, and Bobby Cheng.
\newblock Super tiny language models, 2024.

\bibitem{srivastava2014dropout}
Nitish Srivastava, Geoffrey Hinton, Alex Krizhevsky, Ilya Sutskever, and Ruslan Salakhutdinov.
\newblock Dropout: a simple way to prevent neural networks from overfitting.
\newblock {\em The journal of machine learning research}, 15(1):1929--1958, 2014.

\bibitem{xue2024repeat}
Fuzhao Xue, Yao Fu, Wangchunshu Zhou, Zangwei Zheng, and Yang You.
\newblock To repeat or not to repeat: Insights from scaling llm under token-crisis.
\newblock {\em Advances in Neural Information Processing Systems}, 36, 2024.

\bibitem{radford2019language}
Alec Radford, Jeffrey Wu, Rewon Child, David Luan, Dario Amodei, Ilya Sutskever, et~al.
\newblock Language models are unsupervised multitask learners.
\newblock {\em OpenAI blog}, 1(8):9, 2019.

\bibitem{touvron2023llama}
Hugo Touvron, Louis Martin, Kevin Stone, Peter Albert, Amjad Almahairi, Yasmine Babaei, Nikolay Bashlykov, Soumya Batra, Prajjwal Bhargava, Shruti Bhosale, et~al.
\newblock Llama 2: Open foundation and fine-tuned chat models.
\newblock {\em arXiv preprint arXiv:2307.09288}, 2023.

\bibitem{kaplan2020scaling}
Jared Kaplan, Sam McCandlish, Tom Henighan, Tom~B Brown, Benjamin Chess, Rewon Child, Scott Gray, Alec Radford, Jeffrey Wu, and Dario Amodei.
\newblock Scaling laws for neural language models.
\newblock {\em arXiv preprint arXiv:2001.08361}, 2020.

\bibitem{mukherjee2023orca}
Subhabrata Mukherjee, Arindam Mitra, Ganesh Jawahar, Sahaj Agarwal, Hamid Palangi, and Ahmed Awadallah.
\newblock Orca: Progressive learning from complex explanation traces of gpt-4.
\newblock {\em arXiv preprint arXiv:2306.02707}, 2023.

\bibitem{abdin2024phi}
Marah Abdin, Sam~Ade Jacobs, Ammar~Ahmad Awan, Jyoti Aneja, Ahmed Awadallah, Hany Awadalla, Nguyen Bach, Amit Bahree, Arash Bakhtiari, Harkirat Behl, et~al.
\newblock Phi-3 technical report: A highly capable language model locally on your phone.
\newblock {\em arXiv preprint arXiv:2404.14219}, 2024.

\bibitem{costa2022no}
Marta~R Costa-juss{\`a}, James Cross, Onur {\c{C}}elebi, Maha Elbayad, Kenneth Heafield, Kevin Heffernan, Elahe Kalbassi, Janice Lam, Daniel Licht, Jean Maillard, et~al.
\newblock No language left behind: Scaling human-centered machine translation.
\newblock {\em arXiv preprint arXiv:2207.04672}, 2022.

\bibitem{liu2023dropout}
Zhuang Liu, Zhiqiu Xu, Joseph Jin, Zhiqiang Shen, and Trevor Darrell.
\newblock Dropout reduces underfitting.
\newblock In {\em International Conference on Machine Learning}, pages 22233--22248. PMLR, 2023.

\end{thebibliography}

\appendix
\section{Pre-registration Disclosure}

We include here a copy of the pre-registration clause from our pre-registered report, available here\footnote{\url{https://github.com/LeonGuertler/SuperTinyLanguageModels/blob/main/pre_reports/dropout_prereport.pdf}}:\textit{``This paper is intended to act as a preregistered copy of the eventual research report. Ideally we would enter such a research report blind to partial results, however we have previously run several relevant experiments while developing our framework. Additionally this particular paper is largely recreating the results of Liu et al.~\cite{liu2023dropout} rather than attempting to show particularly new hypotheses, as such this rigour is somewhat less important.
In particular we found dropout to improve modelling performance across the board for simple-wikipedia scale datasets and reduce performance for openwebtext scale datasets (given the 50 million parameter model size), motivating somewhat the experimental design.
''}\\
We largely stuck to this pre-registered version with the following alterations:
\begin{enumerate}
    \item We use Simple English Wikipedia as our smaller dataset for investigating alleviating overfitting with dropout (which we did not specify previously)
    \item For this experiment we use an additional linearly increasing schedule over all $30000$ iterations, which we omitted in the text of the report but not in the image
    \item We conducted additional experiments into the mechanisms of early dropout following a previous experimental design~\cite{liu2023dropout}
\end{enumerate}

\end{document}